\documentclass[conference]{IEEEtran}
\IEEEoverridecommandlockouts
\usepackage{cite}
\usepackage{amsmath,amssymb,amsfonts}
\usepackage{graphicx}
\usepackage{textcomp}
\usepackage{xcolor}
\def\BibTeX{{\rm B\kern-.05em{\sc i\kern-.025em b}\kern-.08em
    T\kern-.1667em\lower.7ex\hbox{E}\kern-.125emX}}

\usepackage{color}

\usepackage{placeins}
\usepackage[hyphens]{url}
\usepackage{hyperref}
\hypersetup{colorlinks=true,breaklinks=true}

\newcommand{\norm}[1]{\left\lVert#1\right\rVert}

\newcommand{\Lagr}{\mathcal{L}}
\DeclareMathOperator{\erf}{erf}

\DeclareMathOperator{\arcsinh}{arsinh}

\usepackage{xr}
\usepackage{booktabs}
\usepackage{algorithm,algpseudocode}
\usepackage{cite}

\usepackage{caption}
\usepackage{bbm}
\usepackage{subcaption}
\usepackage[detect-all]{siunitx}

\usepackage{cleveref}
\usepackage{tikz}
\usepackage{pgfplots}

\usepackage{mathtools}
\usepackage{tabularx} 
\usepackage{todonotes}

\usepackage{mathabx}
\usepackage{mathrsfs}
\usepackage{listings}

\usetikzlibrary{matrix,chains,positioning,decorations.pathreplacing,arrows}
\usetikzlibrary{chains,scopes}

\usepackage{placeins}

\usetikzlibrary{arrows.meta} 
\usepackage[outline]{contour} 
\contourlength{1.4pt}

\tikzset{>=latex} 
\usepackage{xcolor}
\colorlet{myred}{red!80!black}
\colorlet{myblue}{blue!80!black}
\colorlet{mygreen}{green!60!black}
\colorlet{myorange}{orange!70!red!60!black}
\colorlet{mydarkred}{red!30!black}
\colorlet{mydarkblue}{blue!40!black}
\colorlet{mydarkgreen}{green!30!black}
\tikzstyle{node}=[thick,circle,draw=myblue,minimum size=22,inner sep=0.5,outer sep=0.6]
\tikzstyle{node in}=[node,green!20!black,draw=mygreen!30!black,fill=mygreen!25]
\tikzstyle{node hidden}=[node,blue!20!black,draw=myblue!30!black,fill=myblue!20]
\tikzstyle{node convol}=[node,orange!20!black,draw=myorange!30!black,fill=myorange!20]
\tikzstyle{node out}=[node,red!20!black,draw=myred!30!black,fill=myred!20]
\tikzstyle{connect}=[thick,mydarkblue] 
\tikzstyle{connect arrow}=[-{Latex[length=4,width=3.5]},thick,mydarkblue,shorten <=0.5,shorten >=1]
\tikzset{ 
	node 1/.style={node in},
	node 2/.style={node hidden},
	node 3/.style={node out},
}

\usepackage{svg}
\usepackage{xcolor}

\newcommand{\printfnsymbol}[1]{%
	\textsuperscript{\@fnsymbol{#1}}%
} 
\makeatother

\newcommand\copyrighttext{%
	\footnotesize \copyright~2023 IEEE. Personal use of this material is permitted. Permission from IEEE must be obtained for all other uses, in any current or future media, including reprinting/republishing this material for advertising or promotional purposes,creating new collective works, for resale or redistribution to servers or lists, or reuse of any copyrighted component of this work in other works.}
\newcommand\copyrightnotice{%
	\begin{tikzpicture}[remember picture,overlay]
		\node[anchor=south,yshift=10pt] at (current page.south) {\fbox{\parbox{\dimexpr\textwidth-\fboxsep-\fboxrule\relax}{\copyrighttext}}};
	\end{tikzpicture}%
}

\begin{document}

\title{Weight Compander: A Simple Weight Reparameterization for Regularization\\}

\author{\IEEEauthorblockN{Rinor Cakaj}
	\IEEEauthorblockA{\textit{Image Processing} \\
		\textit{Robert Bosch GmbH \& University of Stuttgart}\\
		71229 Leonberg, Germany \\
		Rinor.Cakaj@de.bosch.com}
	\and
	\IEEEauthorblockN{Jens Mehnert}
	\IEEEauthorblockA{\textit{Image Processing} \\
		\textit{Robert Bosch GmbH}\\
		71229 Leonberg, Germany \\
		JensEricMarkus.Mehnert@de.bosch.com}
	\and
	\IEEEauthorblockN{Bin Yang}
	\IEEEauthorblockA{\textit{ISS} \\
		\textit{University of Stuttgart}\\
		70550 Stuttgart, Germany \\
		bin.yang@iss.uni-stuttgart.de}
}

\maketitle

\begin{abstract}
	
Regularization is a set of techniques that are used to improve the
generalization ability of deep neural networks. 
In this paper, we introduce \emph{weight compander} (WC), a novel effective
method to improve generalization by reparameterizing each weight in deep neural
networks using a nonlinear function. It is a general, intuitive, cheap and easy
to implement method, which can be combined with various other regularization
techniques. 
Large weights in deep neural networks are a sign of a more complex network that
is overfitted to the training data. Moreover, regularized networks tend to have
a greater range of weights around zero with fewer weights centered at zero. We
introduce a weight reparameterization function which is applied to each weight
and implicitly reduces overfitting by restricting the magnitude of the weights
while forcing them away from zero at the same time. 
This leads to a more \emph{democratic} decision-making in the network. Firstly,
individual weights cannot have too much influence in the prediction process due
to the restriction of their magnitude. Secondly, more weights are used in the
prediction process, since they are forced away from zero during the training.
This promotes the extraction of more features from the input data and increases
the level of weight redundancy, which makes the network less sensitive to
statistical differences between training and test data. 
From an optimizational point of view, the second effect of WC can be seen as a
reactivation of ``dead'' (near zero) weights to participate in the training.
This increases the probability to find an ensemble of weights which performs
better in the given task. 
We extend our method to learn the hyperparameters of the introduced weight
reparameterization function. This avoids hyperparameter search and gives the
network the opportunity to align the weight reparameterization with the training
progress. 
We show experimentally that using weight compander in addition to standard
regularization methods improves the performance of neural networks. Furthermore,
we empirically analyze the weight distribution with and without weight compander
after training to confirm the companding effects of our method on the weights.
	
\end{abstract}
\copyrightnotice
\section{Introduction}

Deep neural networks contain multiple non-linear hidden layers which make them
powerful machine learning systems \cite{2014_Srivastava}. However, such networks
are prone to overfitting \cite{2013_Wan_CONF} due to the
limited size of training data, the high capacity of the model
and the presence of to many noises in the training data \cite{2019_Ying}.
Overfitting describes the phenomenon where a neural network
(NN) perfectly fits the training data while achieving poor performance on the
test data. Regularization is a set of techniques used to reduce overfitting and
therefore a key element in deep learning \cite{2016_Goodfellow_BOOK}. It allows
the model to generalize well to unseen data. 

Many methods have been developed to regularize NNs, e.g. early stopping
\cite{2007_Yao}, weight penalties as $L_1$-regularization \cite{1996_Tibshirani}
and weight decay \cite{1991_Krogh_CONF}, soft weight sharing \cite{1992_Nowlan},
dropout \cite{2014_Srivastava}, data augmentation  \cite{2019_Shorten} and
ensemble learning methods \cite{1999_Opitz}. 

The architectures of NNs like ResNet \cite{2016_He_CONF}, EfficientNet
\cite{2019_Tan_CONF} or Transformer \cite{2017_Vaswani_CONF} differ widely
across applications, but they typically share the same building blocks, e.g.
weights, neurons and convolutional kernels. 

We take a closer look at the weights of a NN. During the training, the
weights are adjusted to decrease the loss function on the training set. The
success of the optimization procedure depends on the parameterization of the
weights \cite{2016_Salimans_CONF}. The reparameterization of a weight $w \in
\mathbb{R}$ is given by a reparameterization function $\Psi \colon \mathbb{R}
\to \mathbb{R}, \; v \mapsto \Psi(v)$ used to express the weight $w$ in terms of
a new learnable weight $v$ by $w = \Psi(v)$. The idea to use reparameterization
functions has been proposed before to prune or accelerate the training of deep
NNs \cite{2021_Schwarz_CONF, 2016_Salimans_CONF}.

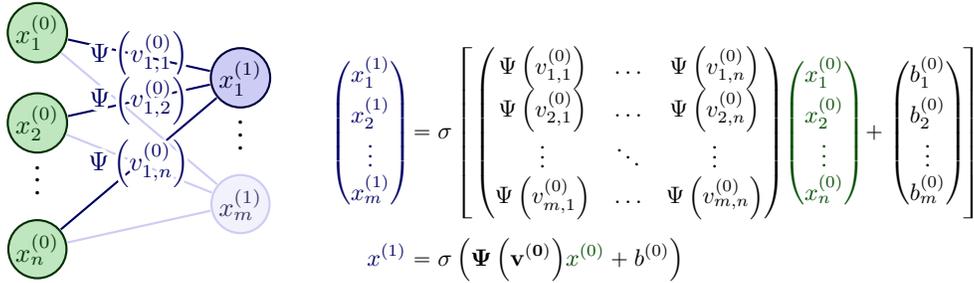
\begin{figure*}[tb]
	\centering
	\begin{tikzpicture}[x=2.7cm,y=1.2cm]
		\message{Neural network activation}
		\def\NI{3} 
		\def\NO{2} 
		\def\yshift{0.4} 
		
		\foreach \i [evaluate={\c=int(\i==\NI); \y=\NI/2-\i-\c*\yshift; \index=(\i<\NI?int(\i):"n");}]
		in {1,...,\NI}{ 
			\node[node in,outer sep=0.6] (NI-\i) at (0,\y) {$x_{\index}^{(0)}$};
		}
		
		\foreach \i [evaluate={\c=int(\i==\NO); \y=\NO/2-\i-\c*\yshift; \index=(\i<\NO?int(\i):"m");}]
		in {\NO,...,1}{ 
			\ifnum\i=1 
			\node[node hidden]
			(NO-\i) at (1,\y) {$x_{\index}^{(1)}$};
			\foreach \j [evaluate={\index=(\j<\NI?int(\j):"n");}] in {1,...,\NI}{ 
				\draw[connect,white,line width=1.2] (NI-\j) -- (NO-\i);
				\draw[connect] (NI-\j) -- (NO-\i)
				node[pos=0.50] {\contour{white}{$\Psi\left(v_{1,\index}^{(0)}\right)$}};
			}
			\else 
			\node[node,blue!20!black!80,draw=myblue!20,fill=myblue!5]
			(NO-\i) at (1,\y) {$x_{\index}^{(1)}$};
			\foreach \j in {1,...,\NI}{ 
				\draw[connect,myblue!20] (NI-\j) -- (NO-\i);
			}
			\fi
		}
		
		\path (NI-\NI) --++ (0,1+\yshift) node[midway,scale=1.2] {$\vdots$};
		\path (NO-\NO) --++ (0,1+\yshift) node[midway,scale=1.2] {$\vdots$};
		
		\def\agr#1{{\color{mydarkgreen}x_{#1}^{(0)}}}
		\node[right,scale=0.9] at (1.4,-0.95)
		{$\begin{aligned}
				{\color{mydarkblue}
					\begin{pmatrix}
						x_{1}^{(1)} \\[0.3em]
						x_{2}^{(1)} \\
						\vdots \\
						x_{m}^{(1)}
				\end{pmatrix}}
				&=
				\color{black}\sigma\left[ \color{black}
				\begin{pmatrix}
					\Psi\left(v_{1,1}^{(0)}\right) & \ldots & \Psi\left(v_{1,n}^{(0)}\right) \\
					\Psi\left(v_{2,1}^{(0)}\right)& \ldots & 
					\Psi\left(v_{2,n}^{(0)}\right) \\
					\vdots  & \ddots & \vdots  \\
					\Psi\left(v_{m,1}^{(0)}\right) & \ldots & 
					\Psi\left(v_{m,n}^{(0)}\right)
				\end{pmatrix}
				{\color{mydarkgreen}
					\begin{pmatrix}
						x_{1}^{(0)} \\[0.3em]
						x_{2}^{(0)} \\
						\vdots \\
						x_{n}^{(0)}
				\end{pmatrix}}
				+
				\begin{pmatrix}
					b_{1}^{(0)} \\[0.3em]
					b_{2}^{(0)} \\
					\vdots \\
					b_{m}^{(0)}
				\end{pmatrix}
				\color{black}\right]\\[0.5em]
				{\color{mydarkblue}x^{(1)}}
				&= \color{black}\sigma\left( \color{black}
				\mathbf{\Psi\left(v^{(0)}\right)} {\color{mydarkgreen}x^{(0)}}+b^{(0)}
				\color{black}\right)
			\end{aligned}$};
	\end{tikzpicture}
	
	\caption{Graphical illustration of the weight compander method applied to a fully connected network. We replace the weights $w$ by $\Psi\left(v\right)$. Throughout the training $v$ is adjusted to decrease the loss function on the training set. \label{graphical_illustration}}
	
\end{figure*}
In this work we introduce \emph{weight compander}, a novel method using a
reparameterization function to regularize general NNs, i.e. to improve their
generalization ability. 

Large weights in deep neural networks are a sign of a more complex network that
is overfitted to the training data. Furthermore, regularized networks tend to
have a greater range of weights around zero with fewer weights centered at zero
\cite{2015_Blundell_CONF}. 

Motivated by these observations, we propose a reparameterization function, which
restricts the magnitude of the weights while forcing them away from zero at the
same time. Therefore, our reparameterization function encourages the model to
\emph{not} rely on a few large weights but rather use many small weights to
benefit from all the available information. 
Thus, our method promotes \emph{weight democracy}, i.e. the influence of
individual weights is limited and more weights are involved in the prediction
process.  

Our reparameterization is a \emph{comp}ressor for weights with large magnitude
and an exp\emph{ander} for weights with small magnitude. Hence, our method is
called \emph{weight compander}.

The nonlinear reparameterization function $\Psi \colon \mathbb{R} \to
\mathbb{R}, \; v \mapsto a \cdot \arctan(\frac{v}{b})$ with hyperparameters
$a,b>0$, which are shared across all weights in the NN, and new learnable weight
$v \in \mathbb{R}$ meets our requirements. The parameter $a$ is used to scale
and restrict the absolute value of the weights $w$ and the parameter $b$ to
squeeze and stretch the function in the $v$ axis. The function is bounded, i.e.
$-\frac{a \pi}{2} < a \cdot\arctan(\frac{v}{b}) < \frac{a \pi}{2}$ for all $v
\in \mathbb{R}$. This restricts the magnitude of the weights. The derivative of
$\Psi(v)$ is relatively larger for small $|v|$ than for large $|v|$. Since the
derivative of $\Psi(v)$ appears in the gradient of $v$, this property is used
during the training to force the weights away from zero.

In summary, weight compander reparameterizes each weight in a NN using the
nonlinear function $\Psi(v)$. Throughout the training $v$ is adjusted to
decrease the loss function on the training set. 
A graphical illustration of a reparameterized fully connected NN is shown in
Figure \ref{graphical_illustration}.

\subsection{Contributions of this Paper and Applications}

In this work we present \emph{weight compander}, a novel method which
reparameterize the weights $w$ to reduce overfitting and which can be applied to
any baseline NN. Our core contributions are:
\begin{itemize}
	\item Defining a reparameterization function $\Psi(v)$ which regularizes NNs.
	\item Analyzing the impact of our method during the training process.
	\item Experimentally showing that using weight compander in addition to
	standard regularization methods increases the performance of
	different models on CIFAR-$10$, CIFAR-$100$, TinyImageNet and ImageNet. We
	improved the test accuracy of ResNet$50$ on CIFAR-$10$ by $0.75\%$, on CIFAR-$100$ by $1.56\%$, on TinyImageNet by $0.86\%$ and on ImageNet by $0.22\%$.
\end{itemize}

\section{Related Work}

\subsection{Regularization}

Regularization is one of the key elements of deep learning
\cite{2016_Goodfellow_BOOK}, allowing the model to generalize well to unseen
data even when trained on a finite training set or with an imperfect
optimization procedure \cite{2017_Kukacka}. There are several techniques to
regularize NNs which can be categorized in groups. Data augmentation
methods like cropping, flipping, and adjusting brightness or sharpness
\cite{2019_Shorten} and cutout \cite{2017_DeVries} transform the training
dataset to avoid overfitting.
Regularization techniques like dropout \cite{2014_Srivastava},
dropblock \cite{2018_Ghiasi_CONF} and dropconnect \cite{2013_Wan_CONF} drop
neurons or weights from the NN during training to prevent units from co-adapting
too much \cite{2014_Srivastava}. Furthermore, NNs can be regularized using
penalty terms in the loss function. Weight decay \cite{1991_Krogh_CONF}
encourages the weights of the NN to be small in magnitude. The
$L_1$-regularization \cite{1996_Tibshirani} forces the weights of
non-relevant features to zero. Normalization techniques \cite{2015_Ioffe_CONF,
	2016_Ba, 2020_Wu} like batch normalization \cite{2015_Ioffe_CONF}, which
normalizes the features by the mean and variance computed within a mini-batch,
act in some cases as regularizers. The resulting stochastic uncertainty of the
batch statistics can benefit generalization \cite{2020_Wu} and in some cases
eliminate the need for dropout \cite{2015_Ioffe_CONF}. Moreover, NNs can be
regularized using early stopping \cite{2007_Yao} and ensemble learning methods
\cite{1999_Opitz}. Weight reparameterization has not been used for
regularization yet. Therefore our introduced method is novel.

\subsection{Weight Reparameterization} \label{sec:weight_reparameterization}

The idea of using reparameterization functions has been
proposed before to prune NNs \cite{2021_Schwarz_CONF} or to accelerate their training
\cite{2016_Salimans_CONF}. Since these reparameterization functions have other
goals, e.g. many low-magnitude weights to easily prune networks, they are not
well suited for regularizing NN.

\subsubsection{Pruning}

Pruning methods are used to reduce the amount of connections in a network
\cite{2015_Han_CONF} with negligible
performance reduction of the network \cite{2020_Wimmer_CONF}. 

Powerpropagation \cite{2021_Schwarz_CONF} is a weight reparameterization method,
which focuses on the inherent effect of gradient based training on model
sparsity. The idea is to reparameterize the weights $w$ of a NN as $ \Psi(v)
= v \cdot |v|^{\alpha - 1}$ for any $\alpha \geq 1$. Due to the chain rule of
calculus, the magnitude of
the parameters raised to $\alpha - 1$ appear in the gradient computation. 
Therefore, smaller magnitude parameters receive smaller gradient updates, while
larger magnitude parameters receive larger updates. The network is then pruned
by removing the low-magnitude weights.

Interspace pruning \cite{2022_Wimmer_CONF} is another pruning method which
reparameterizes convolutional filters as a linear combinations of adaptive
filter basis vectors. Let $h \in \mathbb{R}^{K \times K}$ be a filter with
kernel size $K \times K$, then $h$ can be also modeled in the linear space $\{
\sum_{n=1}^{K^2} \lambda_n \cdot g^{(n)} \colon \, \lambda_n \in \mathbb{R}\}$
where $F \coloneqq \{g^{(1)}, \dots, g^{(K^2)}\} \subset \mathbb{R}^{K \times
	K}$ is the filter basis. This method sets pruned basis coefficients to zero
while training un-pruned basis coefficients and basis vectors jointly, which
leads to a state-of-the-art performance for unstructured pruning methods.

\subsubsection{Accelerate Training of Deep Neural Networks} 

Reducing the training time for a NN with negligible performance reduction is an
important task in deep learning. There are various approaches to
accelerate the training, e.g. normalization methods like batch normalization
\cite{2015_Ioffe_CONF}, progressively freezing layers \cite{2017_Brock_CONF},
prioritizing examples with high loss at each iteration \cite{2019_Jiang} or
dynamically pruning data samples \cite{2021_Raju}.

Weight normalization \cite{2016_Salimans_CONF} is a weight reparameterization
approach that accelerates the convergence of SGD optimization. It
reparameterizes the weight vectors of each layer such that the length of those
weight vectors is decoupled from their direction. In detail, they express the
weight vector $w \in \mathbb{R}^d$ by $w = \Psi(v, g) = \frac{g}{\norm{v}} v$,
where $v \in \mathbb{R}^d$ is the new weight vector and $g \in \mathbb{R}$ the
scalar parameter. In contrast to earlier works \cite{2013_Sutskever_CONF}, they directly perform SGD on the new weight vector $v$ and
scalar parameter $g$.

\section{Weight Compander}

Inspired by the weight reparameterizations for pruning \cite{2021_Schwarz_CONF} and
faster training \cite{2016_Salimans_CONF}, we introduce \textit{weight
compander} to improve the generalization abilities of NNs. We present the
reparameterization function and analyze how established deep learning practices
can be combined with weight compander. Moreover, we extend our method by
learning the new hyperparameters of weight compander with SGD instead of
hyperparameter optimization.

\subsection{Method} \label{Method}

Our aim is to find a function $\Psi(v)$ to reparameterize the weights $w \in
\mathbb{R}$ of a NN to reduce overfitting. In other words, we want to rewrite
the weights $w$ in an equivalent form $w = \Psi(v)$ and optimize the new weights
$v \in \mathbb{R}$ to improve the performance of the NN.

To choose the optimal reparameterization function, we first analyze the effects
of weight decay \cite{1991_Krogh_CONF} and dropout \cite{2014_Srivastava} on the
weights of a NN. Weight decay limits the growth of the weights and
therefore prevents the weights from growing too large \cite{1991_Krogh_CONF}.
Dropout randomly omits hidden units from the network with a probability such
that hidden units cannot rely on other hidden units being present. This prevents units from co-adapting too much
\cite{2014_Srivastava}. Since hidden units are randomly omitted during the
training, the NN cannot rely on a few weights. Therefore, the trained weights
have a greater range around zero with fewer centered at zero
\cite{2015_Blundell_CONF}. 

Hence, we want to find a reparameterization function which strengthens the
regularizing effects of both dropout and weight decay, i.e.
\begin{enumerate}
	\item We want to restrict the magnitude of weights. \label{effect_1}
	\item We want to have a greater range of weights around zero with fewer weights
	centered at zero. \label{effect_2}
\end{enumerate}

Restricting the magnitude of weights restricts the decision-making power of
individual weights. In contrast to weight decay \cite{1991_Krogh_CONF} our
method prevents large weights from the first epoch (see Section
\ref{Experiments}). 

Having a greater range of weights around zero with fewer
weights centered at zero yields an environment where more weights are involved
with more and less equal contributions and more input information are used in
the prediction process. Therefore our method promotes \emph{weight democracy},
i.e. the NN should not rely on a few large weights but rather use many small
weights to benefit from all the available information. In contrast to dropout
\cite{2014_Srivastava}, our method does not have a conflict with batch
normalization \cite{2019_Li_CONF}. Our method does not cause any ``variance
shiftes'' that can harm the performance on DNNs using batch normalization (see
Section \ref{Experiments}). Moreover, our method can be used in addition to
dropout (e.g. in VGGs in Section \ref{Experiments}]), i.e. further increasing
the performance of DNNs.

While the scope of this paper is regularization, from an optimization viewpoint,
the second effect of WC can be seen as reactivating ``dead'' (near zero) weights
to participate in the training, which otherwise would stay near zero during the
training. This increases the probability to find an ensemble of weights which
performs better in the given task. 

At this point one could think that our method contradicts pruning
for model reduction, e.g. \cite{1989_LeCun_CONF}, which claim
that sparse models are superior in the sense of model complexity since they have
a smaller amount of connections while negligible performance reduction of their
networks. In Section \ref{Experiments}, we analyze the weight distribution in
networks with and without WC. Figure \ref{fig:with_weight_repa_zoomed} shows
that nearly $40\%$ of the weights of the first convolutional layer in a trained
ResNet34 are approximately $0$ despite the usage of WC. Moreover, our analysis
shows that the differences in the weight distribution between standard DNNs and
networks using WC were less pronounced in the middle and at the end of the
networks. Hence, WC does not force the network to use \emph{all} weights. It
stimulates the optimization process to use \emph{more} weights in the training,
to find the optimal ensemble of weights. 

Note that the second demand is not in conflict with the aim of the weight decay,
since weight decay only prevents weights from growing too large but does not
force them to zero \cite{1991_Krogh_CONF}. Although the $L_1$-regularization has
a regularizing effect, generally it does not regularize as well as weight decay
\cite{2019_Ma}. The $L_1$-regularization leads to a sparse weight matrix, i.e.
more weights are zero. In contrast to that, weight decay leads to NNs which have
a higher density of weights, i.e. weight decay only shrinks the weights close to
zero, rather than being zero. These networks use more information from
the input data for their prediction, which explains the increased performance
compared to $L_1$-regularization. 

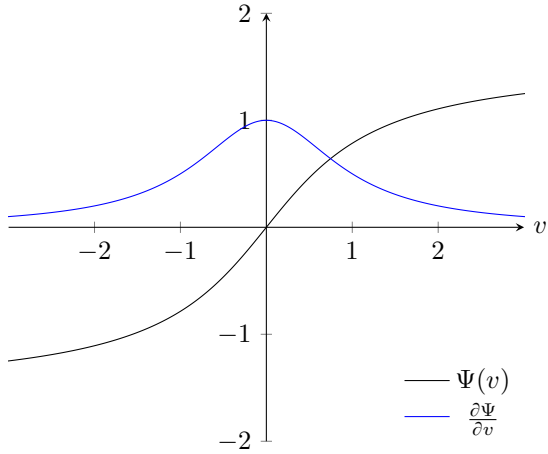
\begin{figure}[tb]
	\centering
	
	\begin{tikzpicture}
		\begin{axis}[
			domain=-10:10,
			xscale=1,yscale=1,
			xmin=-3, xmax=3,
			xtick={-2,-1,1,2},
			ymin=-2, ymax=2,
			samples=1000,
			axis lines=center,
			legend style={at={(1,0)},anchor=south east,draw=none,fill=white,align=left},
			xlabel=$v$,
			every axis x label/.style={
				at={(ticklabel* cs:1.0,-0.1)},
				anchor=west,
			},
			]
			\addplot+[color=black, mark=none] {rad(atan(x)))}; \addlegendentry{$\Psi(v)$};
			\addplot+[color=blue, mark=none] {1/(1+x^2))}; \addlegendentry{$\frac{\partial \Psi}{\partial v}$};
		\end{axis}
	\end{tikzpicture}  
	
	\caption{The reparameterization function $\Psi(v)$ and the derivative of $\Psi(v)$ for $a=1$ and $b=1$, i.e. $\Psi(v)=\arctan(v)$. }\label{Figure_function}
	
\end{figure}

To strengthen the regularizing effects, we introduce the following
reparameterization function:
\begin{align}
	w = \Psi(v) = a \cdot \arctan{\left(\frac{v}{b}\right)}
	\label{reparameterization}
\end{align}
for scalar values $a,b > 0$, which are shared across all weights in the NN.
Figure \ref{Figure_function} shows a plot of $\Psi(v)$ and its derivative. Since
our method targets the weights of a NN, we do not reparameterize batch
normalization layers or biases.

The reparameterization restricts the magnitude of the weights to be in the
interval $(-\frac{a \pi}{2}, \frac{a \pi}{2})$. Hence, the parameter $a$ can be
used to scale \emph{and} restrict the absolute value of the weights. The
gradient of the reparameterized loss $\Lagr(\cdot, \Psi(v))$ w.r.t. $v$ is
\begin{align} \label{computing_gradient}
	\frac{\partial \Lagr (\cdot, \Psi(v))}{\partial v} &= \frac{\partial \Lagr
		(\cdot, \Psi(v))}{\partial \Psi(v)} \cdot \frac{\partial \Psi(v)}{\partial v} \\
	&= \frac{\partial \Lagr (\cdot, \Psi(v))}{\partial \Psi(v)} \cdot \frac{a}{b
		\cdot (1 + \frac{v^2}{b^2})}.
\end{align}

The parameter $b$ can be used to squeeze and stretch the function in the $v$
axis, therefore changing the slope of the function, i.e. the gradient. Note that
both parameters $a$ and $b$ have an effect on the gradient.
In Equation \ref{computing_gradient}, we see that our reparameterization adds
the multiplicative factor $\frac{\partial \Psi(v)}{\partial v}$ to the
derivative $\frac{\partial \Lagr (\cdot, \Psi(v))}{\partial \Psi(v)}$. The
derivative $\frac{\partial \Psi(v)}{\partial v}$ is relatively larger for small
$|v|$ than for large $|v|$ (see Figure \ref{Figure_function}). Thus, weights
near zero are updated relatively stronger than weights farther from zero. This
promotes the weights near zero to have a greater range around zero with fewer
weights centered at zero. 

We now investigate how established deep learning practices can be used with
weight compander. Since we want to compare standard NNs with
reparameterized NNs, the initialization scheme and weight decay regularization
have to be similar to standard NNs.

\subsubsection{Initialization} \label{Initialization}

Two important aspects of reliable training are the initialization of weights
\cite{2015_He_CONF} and normalization layers
such as batch-norm \cite{2015_Ioffe_CONF} or layer-norm \cite{2016_Ba}.
Motivated by \cite{2021_Schwarz_CONF}, we initialize $v$ such that $\Psi(v)$ is
equal to the original initialization of $w$ (e.g. Kaiming initialization
\cite{2015_He_CONF}). Hence, $v$ is initialized by computing the inverse
function of (\ref{reparameterization}):
\begin{align} \label{initialization_phi}
	v = \Psi^{-1}(w) = b \cdot \tan\left(\frac{w}{a}\right).
\end{align}

This ensures that our weight reparameterization maintains all properties of
typical initialization schemes. This initialization method is used in
all experiments in Section \ref{Experiments}.

In general, the weights of standard deep NNs are randomly initialized with small
values, i.e. $ - \frac{\pi}{2} \ll w \ll \frac{\pi}{2}$. 
Generally, Equation \eqref{initialization_phi} is well defined for $a \geq 0.5$
since then $\frac{w}{a}$ is far away from the poles of the
\textit{tangent}. In our experiments, we found out that if $b > 0.5$, then $a <
0.5$ generally harms the performance. In the following, we will therefore assume that $a,b \geq 0.5$. 

\subsubsection{Weight Decay} 

Weight Decay is a regularization method applied to the weights of a NN
\cite{1991_Krogh_CONF}. Let $\mathbf{w}$ be a vector of all weights in a NN.
Weight decay (or $L_2$-regularization) adds a penalty on the $L_2$-norm of the
weights to the loss function, i.e.
\begin{align}
	\bar{\Lagr}(\mathbf{w}) = \Lagr (\mathbf{w}) + \lambda \mathbf{w}^T\mathbf{w},
\end{align}
where $\lambda$ is a value determining the strength of the penalty and
$\mathbf{w}$ a weight vector containing all weights of a NN. To get
comparable results in our experiments we apply weight decay on $w=\Psi(v)$ and
\emph{not} on $v$. 

\subsubsection{Impact of modern optimization Algorithms}

\label{impact_modern_optimiser} Weight compander works fine
with the SGD optimizer, since SGD computes the same gradient step as stated in
Equation \ref{computing_gradient}. However, it is important to analyze the
impact of modern adaptive optimization algorithms \cite{2015_Kingma_CONF,
	2011_Duchi, 2016_Ruder} on our method. The Adam optimizer
\cite{2015_Kingma_CONF} computes moving averages of the gradient and the squared
gradient to obtain an adaptive gradient step that is invariant to re-scaling of
the gradient. Therefore, Adam and other adaptive optimization algorithms
\cite{2015_Kingma_CONF, 2016_Ruder} could reduce the
effect of our method, as weight compander explicitly re-scales the gradient.
To mitigate this issue, we introduce the \emph{modified adaptive optimizer} from
\cite{2021_Schwarz_CONF}. As stated in Section \ref{Method}, we can split the
gradient $\frac{\partial \Lagr}{\partial v}$ in two components $\frac{\partial
	\Lagr }{\partial \Psi(v)}$ and
$\frac{\partial \Psi(v)}{\partial v}$. The modified adaptive optimizer
first computes the adaptive gradient step based on the raw gradient
$\frac{\partial \Lagr}{\partial \Psi(v)}$. Secondly, it multiplies 
$\frac{\partial \Psi(v)}{\partial v}$ to the computed adaptive gradient step to
re-scale it. More details can be found in \cite{2021_Schwarz_CONF}. 

\subsubsection{Inference}

The inference time is the amount of time it takes for a NN to make a prediction
on the test data. The reparameterized NNs have a slightly higher computational
effort, since they need to evaluate the nonlinear weight
reparameterization function for every weight in the NN.

However, this only affects training ($\approx 7\% \textup{--} 14\%$ runtime
overhead depending on the size of the DNN) since we can get rid off the
reparameterization function after the training has finished. 
We can copy the values $w = \Psi(v)$ for optimized $v$ to another network of the
same architecture. Instead of evaluating the function $\Psi(\cdot)$ for every
weight $v$ in the network, we directly save the values $\Psi(v)$ as weights.
This yields a NN with the same inference time as the standard NN. Hence, weight
compander only improves the generalization of NNs during training and has no
negative impact to memory and computational complexity during inference.


\subsection{Learnable Parameters} \label{Learnable_parameters}

Until now we handled the parameters $a$ and $b$ as hyperparameters with fixed
values set before the training. We now present how these parameters
can be learned by SGD. In this sense, $a$ and $b$ become learnable parameters
instead of hyperparameters. Since we do not want to increase the number of
parameters in the NN drastically, we will consider two different versions of
\emph{learnable weight compander}, which differ in their \textit{granularity}.
The \textit{coarse} version shares the parameters $a$ and $b$ across all weights
in the NN. In the \textit{fine} version every layer shares a pair of parameters
$a$ and $b$.

In contrast to the standard weight compander with fixed
hyperparameters $a$ and $b$, we do \emph{not} apply weight decay on $\Psi(v)$.
Assume that weight decay is applied on $\Psi(v)$. The parameters $a$ and $b$ are
learned through SGD, hence they are influenced by weight decay. Since $a$ and
$b$ are shared across one layer or the whole network, weight decay influences
many individual weights at the same time. We got better results applying weight
decay only on the weights $v$ and \emph{not} on the parameters $a$ and $b$.
Sharing the parameters $a$ and $b$ across a layer or across the whole network
implicitly regularizes $a$ and $b$. We will provide empirical results in Section
\ref{Experiments}.

\section{Experiments} \label{Experiments}

We experimentally validate the usefulness of our method in supervised image
recognition, i.e. on CIFAR-$10$/$100$ \cite{2009_Krizhevsky}, on TinyImageNet
\cite{2015_Le_CONF} and on ImageNet \cite{2015_Russakovsky}.
Furthermore, we analyze the differences in the weight distribution
between NNs with and without reparameterization and confirm that weight
compander indeed (i) restricts the magnitude of the weights and (ii) promotes
the weights to have a greater range around zero with fewer weights centered at
zero. We compare NNs using weight compander in addition to
standard regularization methods against NNs using only standard regularization
methods. 

\subsection{Image Classification on CIFAR-$10$/$100$}

We evaluate the performance of WC on the CIFAR-$10$/$100$ classification dataset \cite{2009_Krizhevsky}. The CIFAR-$10$/$100$ dataset
consists of 60,000 $32\times32$ color images, containing 50,000 training images
and 10,000 test images. We used various different network architectures, the
CNN's ResNet$18$, ResNet$34$, ResNet$50$ \cite{2016_He_CONF}, WideResNet-28-10 \cite{2016_Zagoruyko} and VGG$16$/$19$ with BN
\cite{2015_Simonyan_CONF}.

For our experiments we used PyTorch 1.10.1 \cite{2019_Paszke_CONF} and one
Nvidia GeForce 1080Ti GPU. The experiments were run five times with different
random seeds for 200 epochs, resulting in different network initializations,
data orders and additionally in different data augmentations. For every case we
report the mean test accuracy and standard deviation. We used a 9/1-split
between training examples and validation examples and saved the best model on
the validation set. This model was then used for evaluation on the test dataset.

All networks were trained with a batch size of $128$. We used the SGD optimizer
with momentum $0.9$ and initial learning rate $0.1$. For ResNet$18$, ResNet$34$ and ResNet$50$ trained on CIFAR-$10$ we decayed the learning rate by $0.1$ at epoch $90$ and $136$ and used weight decay with the factor $1e-4$ (as in \cite{2016_He_CONF}). For all networks trained on CIFAR-$100$ and for VGG$16$ with BN, VGG19 with BN, we decayed the learning rate by $0.2$ at epoch $60$, $120$ and $160$ and used weight decay with the factor $5e-4$ (as in \cite{2016_Zagoruyko}). For WideResNet-28-10 we used a dropout rate of 0.3. Dropout is also used in the fully connected layers in VGG16 with batch normalization and VGG19 with batch normalization. We want to
point out that we have not done a hyperparameter search for the learning rate,
weight decay, etc. We used the same setting as in \cite{2016_Zagoruyko}.

The ResNet networks are initialized Kaiming-uniform \cite{2015_He_CONF}. In the
VGG networks the linear layers are initialized Kaiming-uniform. The
convolutional layers are initialized with a Gaussian normal distribution with
mean $0$ and standard deviation $\frac{2}{n}$, where $n$ is the size of the
kernel multiplied with the number of output channels in the same layer. In order
to achieve a fair comparison, we initialize the weights $v$ such that the
initialization of $\Psi(v)$ is equal to the initialization of the standard
network, as explained in Section \ref{Initialization}.

For completeness, we compare weight compander with powerpropagation
\cite{2021_Schwarz_CONF} and weight normalization \cite{2016_Salimans_CONF} using
ResNets on CIFAR-$10$/$100$.

\subsubsection{Hyperparameters} \label{Parameters}

For \emph{every} network architecture and \emph{both} datasets we first did a
grid search for the parameters $a$ and $b$ (i.e for $a,b \in \{0.5, 0.6, 0.7,
0.8, 0.9, 1.0\}$) for one seed and picked the values that produced the best
results. Then we conducted the experiments for the other seeds with the
hyperparameters $a$ and $b$ from the first seed. The parameters $a$ and $b$
yielding the best results are shown in Table \ref{tab:optimal_params}. For
WideResNet-28-10 trained on CIFAR-$100$ the parameters that produced the best
results were $a=0.7$ and $b=0.7$. 

\begin{table}[tb]
	\caption{Results of the hyperparameter search for the parameters $a$ and $b$. \label{tab:optimal_params}}
	\begin{tabular*}{\columnwidth}{@{\extracolsep{\fill}}lcc}  
		\toprule
		\textbf{Model} & \textbf{CIFAR10} & \textbf{CIFAR100} \tabularnewline
		\midrule
		ResNet18 & $a=1.0, b=0.6$ & $a=0.6, b=0.6$ \tabularnewline
		ResNet34 & $a=0.8, b=0.5$ & $a=0.7, b=0.8$ \tabularnewline
		ResNet50 & $a=1.0, b=0.6$ & $a=1.0, b=0.8$ \tabularnewline
		VGG16 & $a=0.7, b=0.7$ & $a=0.7, b=1.0$ \tabularnewline
		VGG19 & $a=0.5, b=0.7$ & $a=0.7, b=0.8$ \tabularnewline
		\bottomrule
	\end{tabular*} 
\end{table} 

\subsubsection{Results}

Table \ref{tab:cifar10} and \ref{tab:cifar100} present the results of our
experiments on CIFAR-$10$ and CIFAR-$100$. The additions to the names of the models describe the reparameterization method, i.e. PP = Powerpropagation, WN = Weight Normalization, WC = Weight Compander. Our method improves the accuracy for all ResNets and
VGGs. The additional gain in performance by reparameterizing the weights in the
ResNet50 network are worth noting, i.e. $+0.75\%$ on CIFAR-$10$ and
$+1.56\%$ on CIFAR-$100$. We observe that standard ResNet networks
tend to generalize worse if more parameters are trainable. This can be
counteracted by using weight compander.

Powerpropagation \cite{2021_Schwarz_CONF} allows parameters with larger magnitudes to
adapt, while smaller magnitude parameters are restricted. Hence, low-magnitude
parameters are largely unaffected by learning. This strong restriction lead to a
decrease of the model performance.

Weight normalization \cite{2016_Salimans_CONF} improves the conditioning of the
optimization problem and speeds up convergence of SGD \cite{2016_Salimans_CONF}.
This slightly increases the performance compared to the baseline networks.
However, weight compander outperforms weight normalization (except for
ResNet$18$ on CIFAR-$10$). Especially for deep networks (ResNet$34$, ResNet$50$) trained on more complex tasks (CIFAR-$100$) weight compander performs much better than weight normalization.
	
\begin{table}[tb]
	\caption{Accuracy of experiments on CIFAR-$10$. \label{tab:cifar10}}
	\begin{tabular*}{\columnwidth}{@{\extracolsep{\fill}}lc} 
		\toprule
		\textbf{Model} & \textbf{Accuracy} \tabularnewline
		\midrule
		ResNet18  & $94.04\% \pm 0.08\%$ \tabularnewline
		ResNet18 + PP & $92.94\% \pm 0.17\%$ \tabularnewline
		ResNet18 + WN  & $94.26\% \pm 0.14\%$ \tabularnewline
		ResNet18 + WC & $\mathbf{94.44\% \pm 0.12\%}$ \tabularnewline
		\midrule
		ResNet34  & $93.69\% \pm 0.30\%$ \tabularnewline
		ResNet34 + PP  & $93.29\% \pm 0.16\%$ \tabularnewline
		ResNet34 + WN  & $94.17\% \pm 0.22\%$ \tabularnewline
		ResNet34 + WC & $\mathbf{94.43\% \pm 0.25\%}$ \tabularnewline
		\midrule
		ResNet50  & $93.31\% \pm 0.36\%$  \tabularnewline
		ResNet50 + PP & $93.47\% \pm 0.14\%$ \tabularnewline
		ResNet50 + WN & $93.89\% \pm 0.23\%$ \tabularnewline
		ResNet50 + WC & $\mathbf{94.06\% \pm 0.27\%}$ \tabularnewline
		\midrule
		VGG16-BN & $93.27\% \pm 0.12\%$ \tabularnewline
		VGG16-BN + WC & $\mathbf{93.52\% \pm 0.19\%}$ \tabularnewline
		VGG19-BN  & $93.21\% \pm 0.08\%$  \tabularnewline   
		VGG19-BN + WC & $\mathbf{93.33\% \pm 0.25\%}$  \tabularnewline
		\bottomrule
	\end{tabular*}
\end{table}

\begin{table}[tb]
	\caption{Accuracy of experiments on CIFAR-$100$.  \label{tab:cifar100}}
	\begin{tabular*}{\columnwidth}{@{\extracolsep{\fill}}lc} 
		\toprule
		\textbf{Model} & \textbf{Accuracy} \tabularnewline
		\midrule
		ResNet18 & $76.47\% \pm 0.20\%$ \tabularnewline
		ResNet18 + PP & $74.70\% \pm 1.32\%$ \tabularnewline
		ResNet18 + WN & $\mathbf{76.83\% \pm 0.22\%}$  \tabularnewline
		ResNet18 + WC & $76.73\% \pm 0.09\%$ \tabularnewline
		\midrule
		ResNet34 & $77.07\% \pm 0.45\%$ \tabularnewline
		ResNet34 + PP  & $74.70\% \pm 1.32\%$ \tabularnewline
		ResNet34 + WN  & $76.98\% \pm 0.39\%$ \tabularnewline
		ResNet34 + WC & $\mathbf{77.54\% \pm 0.42\%}$ \tabularnewline
		\midrule
		ResNet50 & $76.17\% \pm 0.69\%$ \tabularnewline
		ResNet50 + PP & $75.23\% \pm 0.25\%$  \tabularnewline
		ResNet50 + WN & $76.10\% \pm 0.25\%$ \tabularnewline
		ResNet50 + WC & $\mathbf{77.73\% \pm 0.40\%}$ \tabularnewline
		\midrule
		WRN-28-10 & $80.09\% \pm 0.14\%$\tabularnewline
		WRN-28-10 + WC & $\mathbf{80.24\% \pm 0.36\%}$ \tabularnewline
		\midrule
		VGG16-BN  & $72.48\% \pm 0.35\%$ \tabularnewline
		VGG16-BN + WC & $\mathbf{72.74\% \pm 0.25\%}$ \tabularnewline
		VGG19-BN & $71.34\% \pm 0.14\%$ \tabularnewline
		VGG19-BN + WC & $\mathbf{71.86\% \pm 0.26\%}$ \tabularnewline
		\bottomrule
	\end{tabular*} 
\end{table}

\subsection{Image Classification on TinyImageNet}

TinyImageNet \cite{2015_Le_CONF} contains 100,000 images of 200 classes (500 for
each class) downsized to $64\times64$ colored images. Each class has 500
training images and 50 validation images.

For our experiments on TinyImageNet, we used PyTorch 1.10.1
\cite{2019_Paszke_CONF} and one Nvidia GeForce 1080Ti GPU. The experiments were
run three times with different random seeds for 300 epochs using a ResNet$50$.
Following the common practice, we report the top-1 classification accuracy on
the validation set. Analogously, to the experiments on CIFAR, we determined the
optimal hyperparameters $a=0.7$ and $b=0.8$ for ResNet$50$. We trained all the models using the
SGD optimizer with momentum $0.9$ and initial learning rate $0.1$ decayed by
factor 0.1 at epochs 75, 150, and 225. Moreover, we used weight decay with the
factor $1e-4$. The batch size is set to 128. The training data is augmented by
using random crop with size $224$, random horizontal flip and normalization. The
validation data is augmented by resizing to $256\times256$, using center crop
with size $224$ and normalization.

\subsubsection{Results}

Table \ref{tab:TinyImageNet} presents the top-1 accuracy of our experiments on
TinyImageNet. Using WC increases the accuracy of ResNet50 on TinyImageNet by $0.86\%$.

\begin{table}[tb]
	\caption{Top-1 Accuracy of experiments on TinyImageNet. \label{tab:TinyImageNet}}
	\begin{tabular*}{\columnwidth}{@{\extracolsep{\fill}}lc} 
		\toprule
		\textbf{Model} & \textbf{Accuracy} \tabularnewline
		\midrule
		ResNet50 & $65.08\% \pm 0.38\%$ \tabularnewline
		ResNet50 + WC & $\mathbf{65.93\% \pm 0.12\%}$ \tabularnewline
		\bottomrule
	\end{tabular*} 
\end{table}

\subsection{ImageNet}

The ILSVRC 2012 classification dataset \cite{2015_Russakovsky} contains 1.2
million training images, 50,000 validation images, and 150,000 testing images.
Images are labeled with 1,000 categories. 

For our experiments on ImageNet, we used PyTorch 1.10.1 \cite{2019_Paszke_CONF}
and 4 Nvidia GeForce 1080Ti GPU. The experiments were run three times
with different random seeds for 300 epochs using ResNet$50$. Following the
common practice, we report the top-1 classification accuracy on the validation
set. Analogously, to the experiments on CIFAR and TinyImageNet, we determined
the optimal hyperparameters $a=0.6$ and $b=0.7$. We trained all the
models using the SGD optimizer with momentum $0.9$ and  initial learning rate
0.1 decayed by factor 0.1 at epochs 75, 150, 225. Moreover, we used weight decay
with the factor $1e-4$. The batch size is set to 256. The training data and
validation data is augmented as described for TinyImageNet.

\subsubsection{Results}

Table \ref{tab:ImageNet} presents the top-1 accuracy of our experiments on
ImageNet. Using WC increases the accuracy of ResNet50 on ImageNet by $0.22\%$.

\begin{table}[tb]
	\caption{Top-1 Accuracy of experiments on ImageNet. \label{tab:ImageNet}}
	\begin{tabular*}{\columnwidth}{@{\extracolsep{\fill}}lc} 
		\toprule
		\textbf{Model} & \textbf{Accuracy} \tabularnewline
		\midrule
		ResNet50 & $76.80\% \pm 0.11\%$ \tabularnewline
		ResNet50 + WC & $\mathbf{77.02\% \pm 0.08\%}$ \tabularnewline
		\bottomrule
	\end{tabular*} 
\end{table}

\subsection{Weight Distribution} \label{weight_distribution}

In the following, we analyze the effect of weight compander on the weight
distribution of the networks. We experimentally confirm that weight compander
strengthens the two phenomenons mentioned in Section \ref{Method}, i.e. (i) it
restricts the magnitude of the weights and (ii) it forces the network to have a
greater range of weights around zero with fewer weights centered at zero. 

\begin{figure*}[tb]
	\centering
	\begin{subfigure}{.5\textwidth}
		\centering		\includegraphics[width=\linewidth]{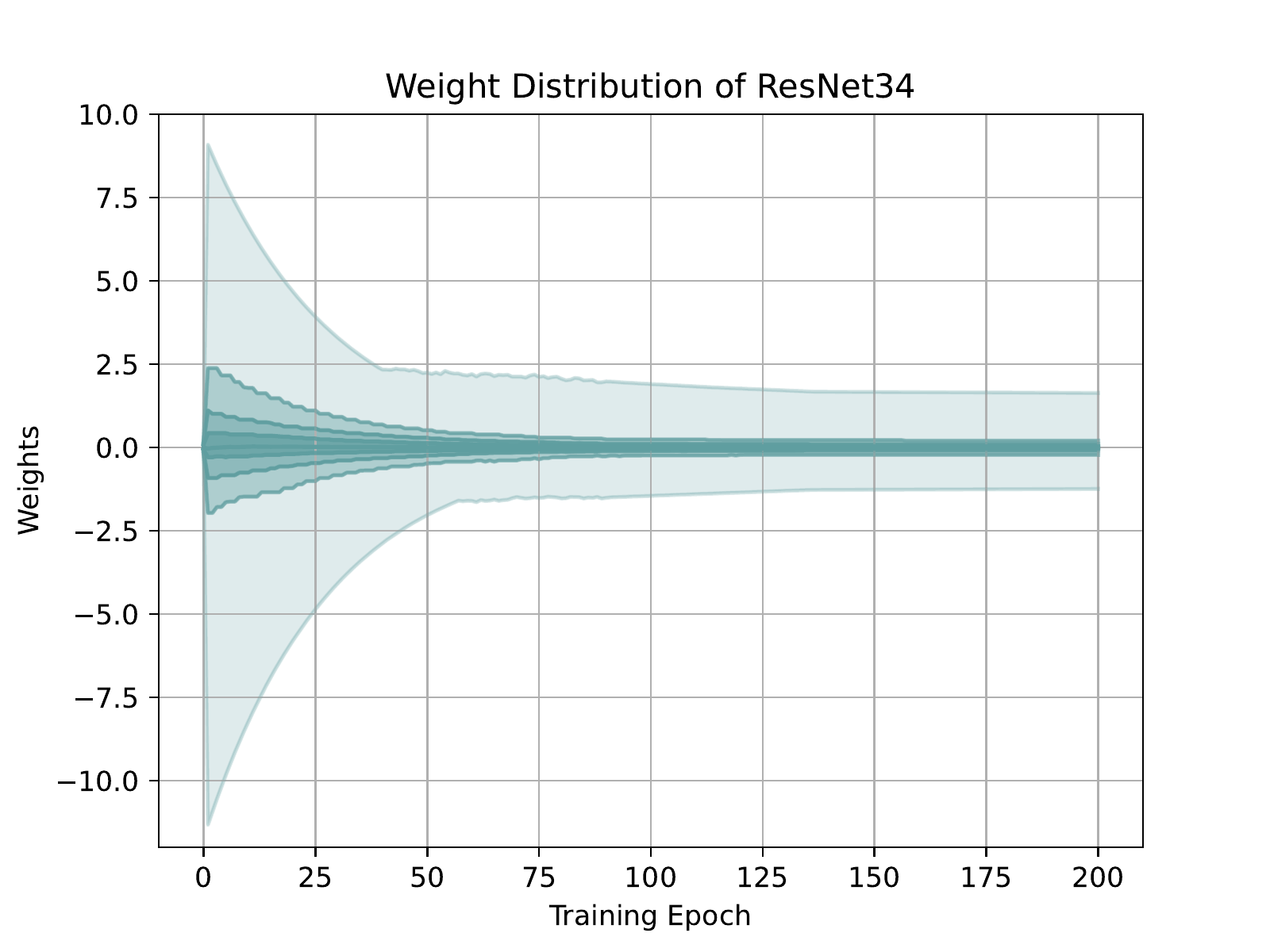}
		\caption{Without weight compander}
		\label{fig:without_weight_repa}
	\end{subfigure}%
	\begin{subfigure}{.5\textwidth}
		\centering
		\includegraphics[width=\linewidth]{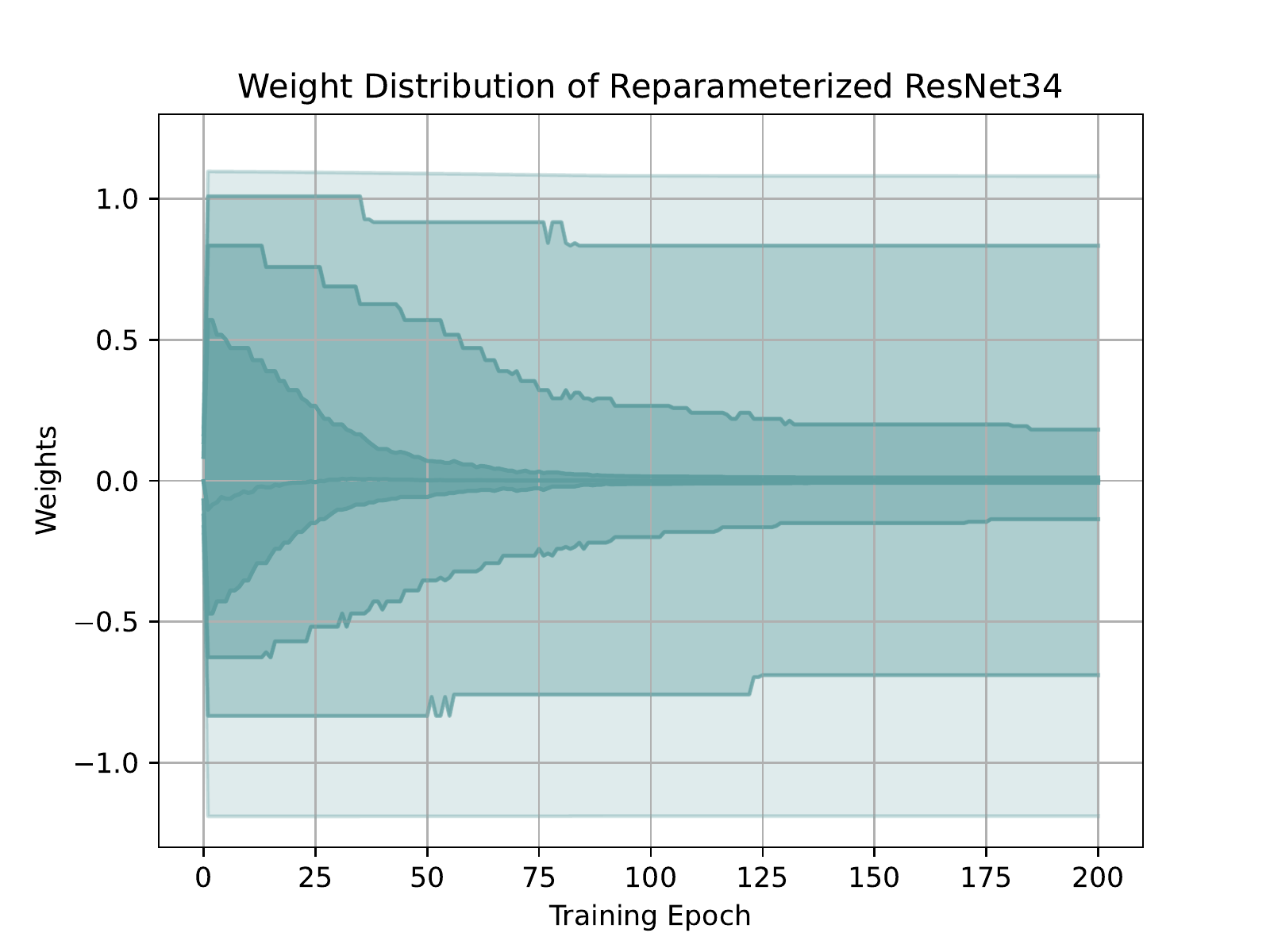}
		\caption{With weight compander}
		\label{fig:with_weight_repa}
	\end{subfigure}
	\caption{Weight distribution of $w$ per epoch in the first convolutional layer of ResNet34 trained on CIFAR10 without weight compander \ref{fig:without_weight_repa}
		and with weight compander \ref{fig:with_weight_repa}. From top to
		bottom, the lines represent percentiles with values: $\text{maximum}, 93\%,
		84\%, 69\%, 50\%, 31\%, 16\%, 7\%, \text{minimum}$.}
	\label{fig:distribution}
\end{figure*}

Setting $a=0.8, b=0.5$ produced the best results for ResNet34 on CIFAR-$10$
using weight compander in addition to standard regularization methods. 
Figure \ref{fig:distribution} shows the weight distribution of the first
convolutional layer of a ResNet34 trained on CIFAR-$10$ per epoch with and
without weight compander. Note that both NNs have the same weight distribution
at epoch $0$. 
Figure \ref{fig:without_weight_repa} shows that the weights in a standard neural
network can grow very large at the beginning of the training. Weight decay does
not play an important role at the beginning of the optimization procedure due to
the high training loss and therefore does not prevent the weights to grow very
large in the first epochs. Figure \ref{fig:with_weight_repa} illustrates that
weight compander bounds the maximal and minimal value of the
weights and therefore forces the weights to stay in the interval $(-\frac{a
\pi}{2}, \frac{a \pi}{2})$ during the training. Since each weight has a small
magnitude, the influence of individual weighs in the prediction process is
limited. 

Standard NNs generally do not have very large weights at the end of the
training. Since we restrict the magnitude of the weights from the beginning of
the training, our method forces SGD to search from the first epoch for a
solution with smaller weights. This also explains the increased performance of
our method.

\begin{figure*}[tb]
	\centering
	\begin{subfigure}{.5\textwidth}
		\centering
		\includegraphics[width=\linewidth]{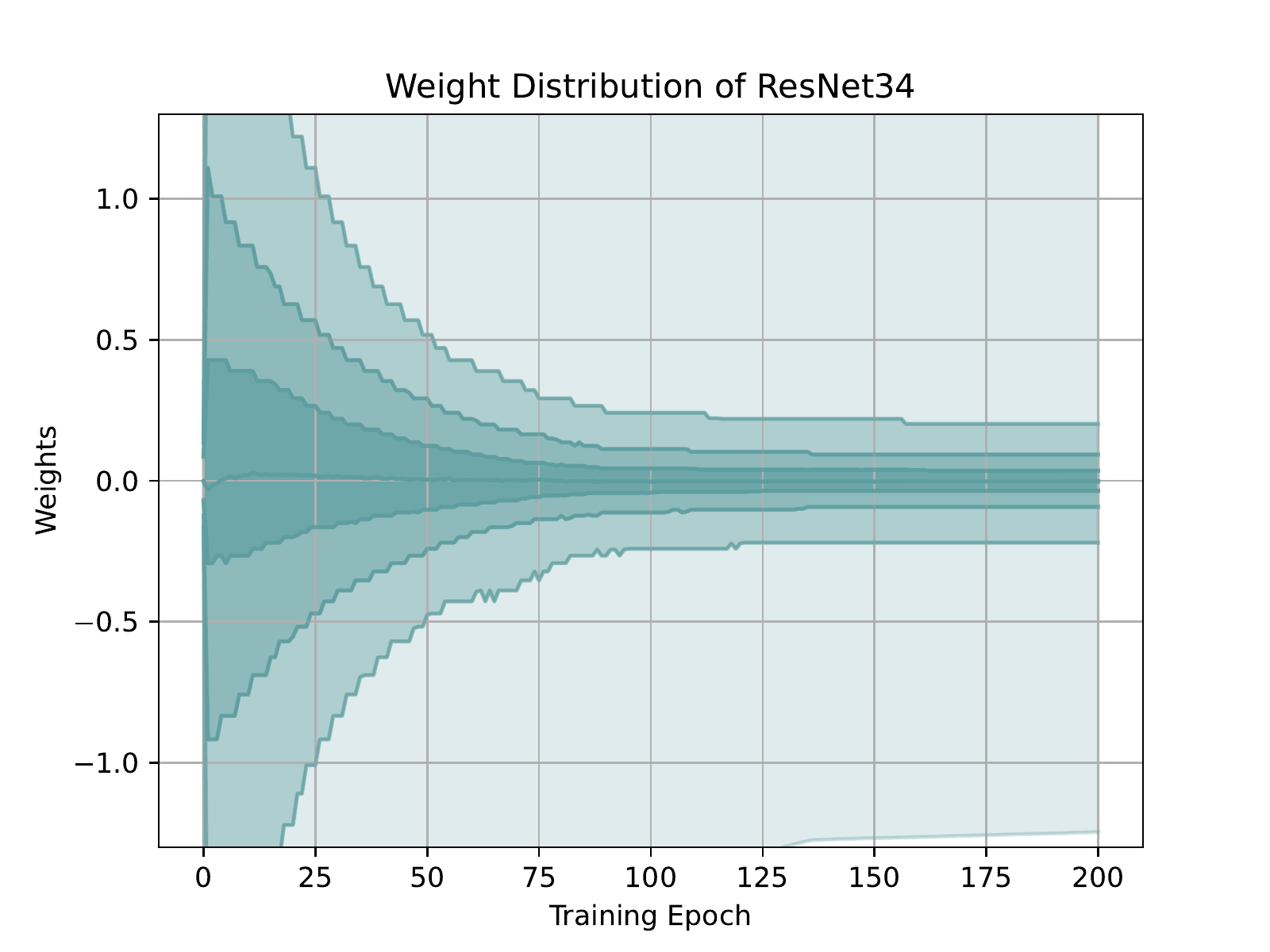}
		\caption{Without weight compander}
		\label{fig:without_weight_repa_zoomed}
	\end{subfigure}%
	\begin{subfigure}{.5\textwidth}
		\centering
		\includegraphics[width=\linewidth]{Figures/ResNet34_repa_conv1_labeled}
		\caption{With weight compander}
		\label{fig:with_weight_repa_zoomed}
	\end{subfigure}
	\caption{Weight distribution of $w$ per epoch in the first convolutional layer of ResNet34 on CIFAR10 without weight compander \ref{fig:without_weight_repa_zoomed} (zoomed view of Figure \ref{fig:without_weight_repa}) and with weight compander \ref{fig:with_weight_repa_zoomed}.}
	\label{fig:distribution_zoomed}
\end{figure*}

Figure \ref{fig:distribution_zoomed} shows that weight compander encourages the
weights to spread around zero, i.e. to have a greater range of weights around
zero with fewer weights centered at zero.
Therefore, more weights are involved in the prediction process. Due to the
increased level of weight redundancy, the model is less sensitive to statistical
differences between training and test data. 

The weight distributions show that weight compander promotes \emph{weight
democracy} in NNs, i.e. the NN does not rely on a few large weights but rather
use many small weights to benefit from all the available information.

The differences in weight distribution were less pronounced in the middle and at
the end of the networks. Early convolutional layers detect low-level features
such as edges and curves, while filters in higher layers are learned to encode
more abstract features. Hence, reparameterized deep NNs extract
more information from the input data than standard NNs. Nevertheless, each layer
must be reparameterized to achieve the best
performance.

\subsection{Learnable Parameters} \label{Learnable_params}


Since a hyperparameter search takes a lot of time and can be computationally
extensive, we analyze the performance of \emph{learnable weight
regulation} where the parameters $a$ and $b$ are learned by SGD. The experiments
were done in the same setting as described at the beginning of Section \ref{Experiments}. We did a small hyperparameter search
to find the optimal initialization values for $a$ and $b$, i.e.
$a,b \in \{0.5, 1.0\}$. As described in Section \ref{Learnable_parameters}, we
compare two versions of learnable weight compander, i.e. the \textit{fine}
version where the parameters $a$ and $b$ are shared across each layer and the
\textit{coarse} version where the parameters are shared across all weights in
the NN. However, the \textit{fine} version performed significantly better
than the \textit{coarse} version. This implies that different layers in the
network need different reparameterizations. Therefore, we only present the
results of the \textit{fine} version. 

Initializing the \textit{fine} version with $a=0.5$ and $b=1.0$ yields the best
performance. Table \ref{tab:learnable_params} shows the results for ResNets
trained on CIFAR-$10$.
\begin{table}[tb]
	\caption{Accuracy of learnable weight compander on CIFAR-$10$ for the \textit{fine} version. \label{tab:learnable_params}}
	\begin{tabular*}{\columnwidth}{@{\extracolsep{\fill}}lc} 
		\toprule
		\textbf{Model} & \textbf{Accuracy} \tabularnewline
		\midrule
		ResNet18 & $94.04\% \pm 0.08\%$ \tabularnewline
		ResNet18 + WC & $\mathbf{94.16\% \pm 0.08\%}$ \tabularnewline
		ResNet34 & $93.69\% \pm 0.30\%$ \tabularnewline
		ResNet34 + WC & $\mathbf{94.46\% \pm 0.07\%}$ \tabularnewline
		ResNet50 & $93.31\% \pm 0.36\%$ \tabularnewline
		ResNet50 + WC & $\mathbf{94.80\% \pm 0.14\%}$ \tabularnewline
		\bottomrule
	\end{tabular*} 
\end{table}
In contrast to weight compander with fixed hyperparameters $a$ and $b$,
learnable weight compander could achieve significantly better performance on
ResNet50. 

Training NNs requires minimizing a high-dimensional non-convex loss function.
Different network architectures affect the loss landscape and therefore the
generalization error and trainability \cite{2018_Li_CONF}. In the case of
learnable weight compander, SGD has the opportunity to change the
reparameterization of each layer during the training. Therefore the loss
landscape changes during the training. We think that learnable weight compander
performs better since the optimization procedure has now the opportunity to
affect the loss landscape. The performance on ResNet18 was
slightly worse compared to the case with fixed hyperparameters. These results
could be improved by finding the optimal hyperparameters for the learnable version.

We logged the values of the parameters $a$ and $b$ for each layer to understand
their impact on the performance.  
The parameters $a$ in the convolutional layers have not changed much compared to
the linear layer. They stayed close to their initial values or moved in a
neighborhood around their initial value ($\pm 0.35$). Interestingly, the
parameter $a$ in the
last linear layer had a very unusual development. Their absolute values were
very high, e.g. $8.5$ for ResNet18 on one seed. Similar or even bigger values
for $a$ were observed on ResNet34 and ResNet50.

The parameters $b$ have changed more in
the optimization process. For the convolutional layers they were mostly in the
interval $[0.5, 2.5]$ at the end of the training. Interestingly, the values for
the parameters $b$ were higher at earlier layers than at later layers in the
network. Again, the linear layer is an exception where the parameters $b$ took
values from $0.5$ to $11.15$ over the five seeds. However, the network
automatically
converges to the desired distribution of weights described in the previous
sections. For some layers the differences of the weight distribution were less
pronounced.

\subsection{Comparison with other $S$-shaped function} \label{results_experiments_other_function}

Additionally to our reparameterization function presented in the paper, we
investigated other odd $S$-shaped functions, i.e. functions that have a
rotational symmetry with respect to the origin. This means that their graph
remains unchanged after rotation of $180$ degrees about the origin. Note that
it is not possible to use an arbitrary reparameterization function, e.g. using
$w = \Psi(v) = v^2$ restricts the weights to be only positive and the
derivative of $v^2$ grows for growing $|v|$. These effects harm the optimization
process.

We had a look at the following other reparameterization functions $ w = \Psi(v)
= a \cdot \arcsinh{\left(\frac{v}{b}\right)} $ and  $ w = \Psi(v) = a \cdot
\erf{\left(\frac{v}{b}\right)} $ for $a,b > 0$. In contrast to the $\arctan$ and
$\erf$, $\arcsinh$ is unbounded.
We included it in our investigations since it is similar to $S$-shaped curves in
the neighborhood around zero. Analogous to the experiments in Section
\ref{Experiments} in the paper, we did a
grid search for every network and dataset for the parameters $a$ and $b$ (i.e
for $a, b \in \{0.5, \dots, 1.0\}$) for one seed and picked the values that
produced the best results. Then we conducted the other experiments for
the other seeds with the hyperparameters $a$ and $b$ from the first seed. We
used the same settings as presented in Section \ref{Experiments}.

Table \ref{tab:results_cifar10_cifar100_s_shaped} present the results of ResNets
trained on CIFAR-$10$ and CIFAR-$100$, which are reparameterized using Equation
$\arcsinh$ and Equation
$\erf$. Reparameterizing ResNets with
$\arcsinh$ or $\erf$ improves the accuracy for
all ResNets compared to the baseline. However, reparameterizing the NNs with
$\arcsinh$ performs worse than reparameterizing the networks with $\arctan$
except for ResNet34 trained on CIFAR-$10$.

\begin{table}[tb]
	\caption{Results of ResNets trained on CIFAR-$10$ and CIFAR-$100$ reparameterized with $\arcsinh$ and $\erf$. \label{tab:results_cifar10_cifar100_s_shaped}}
	\begin{tabularx}{\columnwidth}{@{\extracolsep{\fill}} lcc} 
		\toprule
		\textbf{Model} & \textbf{Accuracy CIFAR-$10$} & \textbf{Accuracy CIFAR-$100$} \tabularnewline
		\midrule
		ResNet18\_arcsinh & $94.34\% \pm 0.24\%$ & $76.59\% \pm 0.25\%$ \tabularnewline
		ResNet18\_erf & $94.36\% \pm 0.26\%$ & $\mathbf{76.80\% \pm 0.11\%}$\tabularnewline
		ResNet18 + WC & $\mathbf{94.44\% \pm 0.12\%}$ & $76.73\% \pm 0.09\%$ \tabularnewline
		\midrule
		ResNet34\_arcsinh & $\mathbf{94.68\% \pm 0.39\%}$ & $77.21\% \pm 0.23\%$ \tabularnewline
		ResNet34\_erf & $94.52\% \pm 0.14\%$ & $77.26\% \pm 0.39\%$ \tabularnewline
		ResNet34 + WC  & $94.43\% \pm 0.25\%$ & $\mathbf{77.54\% \pm 0.42\%}$  \tabularnewline
		\midrule
		ResNet50\_arcsinh & $93.82\% \pm 0.34\%$ & $76.76\% \pm 0.42\%$ \tabularnewline
		ResNet50\_erf & $93.87\% \pm 0.30\%$ & $77.51\% \pm 0.37\%$ \tabularnewline
		ResNet50 + WC & $\mathbf{94.06\% \pm 0.27\%}$ & $\mathbf{77.73\% \pm 0.40\%}$ \tabularnewline
		\bottomrule
	\end{tabularx} 
\end{table}

\subsection{Vanishing Gradients}

We did not have vanishing gradients using WC. Our method is applied to the
weights. Since most of the weights are distributed around zero (see Figure
\ref{fig:with_weight_repa_zoomed}), the gradient of the reparameterization
function is for the most weights unequal to zero. Moreover, for some cases ($a >
b$) the derivative of the reparameterization function evaluated at $0$ is even
greater than $1$. It is important to note that our nonlinear
\emph{reparameterization} function should not be compared with nonlinear
\emph{activation} functions.

\section{Conclusion and Future Work}

We presented weight compander, a simple reparameterization of the weights in a
NN to reduce overfitting. It is based on two premises (i) large weights are a
sign of a more complex network that is overfitted to the training data and (ii)
regularized networks tend to have a greater range of weights around zero with
fewer weights centered at zero. Our proposed reparameterization function
implicitly supports the network to restrict the magnitude of weights while
forcing them away from zero at the same time. Therefore, our method promotes
\emph{weight democracy}, i.e. the influence of individual weights is limited and
more weights are involved in the prediction process. 

Using weight compander in addition to commonly used
regularization methods (i.e. \cite{2019_Shorten,2015_Ioffe_CONF,1991_Krogh_CONF,
2019_Loshchilov_CONF, 2014_Srivastava}) increased the performance of ResNets on
CIFAR-$10$, CIFAR-$100$, TinyImageNet and ImageNet and of VGGs on CIFAR-$10$ and
CIFAR-$100$. We improved the test accuracy of ResNet50 on CIFAR-$100$ by
$1.56\%$, on TinyImageNet by $0.86\%$ and on ImageNet by $0.22\%$.
Furthermore, we analyzed the effect of weight compander
on the weight distribution. Finally, we presented learnable weight compander
where the introduced hyperparameter are learned by SGD.

We have not explored the full range of possibilities of weight compander. Our
future work will include analyzing the role of different reparameterization
functions and further developing learnable weight compander. We did experiments
with $\arcsinh$ and $\erf$ where we could achieve similar performance boosts as
for $\mathit{\arctan}$. We will analysis the optimal weight distributions in
DNNs to increase the performance gains of weight reparameterization methods.
Moreover, we will investigate how weight reparameterization of DNNs can improve
other tasks in deep learning. An extension of our introduced method could be to
make the reparameterization more complex, i.e to have a correlation between
different weights.

\bibliographystyle{IEEEtran}
\bibliography{Reparametrization}

\end{document}